%% file: main.tex
\documentclass[11pt,a4paper]{article}
\usepackage[hyperref]{acl2020}
\usepackage{times}
\usepackage{latexsym}

\newcommand{\eat}[1]{}

\newcommand{\blue}[1]{\textcolor{blue}{#1}}

\usepackage{quoting}
\newenvironment{myquote}{                   
  \parskip 0mm \begin{quoting}[vskip=0mm,leftmargin=2mm]}{
\end{quoting}}
\newenvironment{myquote2}{                   
  \parskip 0mm \begin{quoting}[vskip=0mm,leftmargin=10mm]}{
\end{quoting}}
\newenvironment{ite}{                     
     \parskip 0cm \begin{itemize} \parskip 0cm \parsep 0cm \itemsep 0cm \topsep 0cm}{
        \end{itemize}} 
\newenvironment{des}{                 
     \parskip 0cm \begin{list}{}{\parsep 0cm \itemsep 0cm \topsep 0cm}}{
       \end{list}} 

\usepackage{microtype}

\aclfinalcopy 

\usepackage{graphicx}
\usepackage{subcaption}
\usepackage{amsmath}
\usepackage{mathtools}
\usepackage{amsfonts}

\newcommand{\cut}[1]{}

\newcommand{\GenericsKB}{\textsc{GenericsKB}}
\newcommand{\GenericsKBBest}{\textsc{GenericsKB-Best}}

\title{GenericsKB: A Knowledge Base of Generic Statements}

\author{Sumithra Bhakthavatsalam, Chloe Anastasiades, Peter Clark \\
        Allen Institute for Artificial Intelligence, Seattle, WA \\
        {\tt \{sumithrab,chloea,peterc\}@allenai.org}
        }

\date{}

\begin{document}

\maketitle


\input{paper.tex}


\bibliography{references}
\bibliographystyle{acl_natbib}



\input{appendix}

\end{document}

%% file: paper.tex
\begin{abstract}
  We present a new resource for the NLP community, namely a large (3.5M+ sentence) knowledge base of {\it generic statements},
  e.g., ``Trees remove carbon dioxide from the atmosphere'', collected from multiple corpora.
  This is the first large resource to contain {\it naturally occurring} generic sentences,
  as opposed to extracted or crowdsourced triples, and thus is rich in high-quality,
  general, semantically complete statements.
  All \GenericsKB~sentences are annotated with their topical term,
  surrounding context (sentences), and a (learned) confidence.
  We also release \GenericsKBBest~(1M+ sentences), containing the best-quality generics in
  \GenericsKB~augmented with selected, synthesized generics from WordNet and ConceptNet.
  In tests on two existing datasets requiring multihop reasoning (OBQA and QASC), we find using
  \GenericsKB~can result in higher scores and better explanations than using a much larger corpus.
  This demonstrates that \GenericsKB~can be a useful resource for NLP applications, as well 
  as providing data for linguistic studies of generics and their semantics.\footnote{
    \GenericsKB~is available at https://allenai.org/data/genericskb}
\end{abstract}

\section{Introduction}

While deep learning systems have achieved remarkable performance
trained on general text, NLP researchers frequently seek out
additional repositories of general/commonsense knowledge to
boost performance further, e.g., 
\cite{Icarte2017HowAG,Wang2018ModelingSP,Yang2019EnhancingPL,Peters2019KnowledgeEC,Liu2019KBERTEL,Paul2019RankingAS}.
However, there are only a limited number of repositories currently available,
with ConceptNet \cite{speer2017conceptnet} and WordNet \cite{fellbaum1998wordnet} being popular choices.
In this work we contribute a new, novel resource, namely a large collection
of contextualized {\it generic sentences}, as an
additional source of general knowledge, and to
help fill gaps with existing repositories.
The resource, called \GenericsKB, is the first to contain {\it naturally occurring} generic sentences,
as opposed to extracted or crowdsourced triples, and thus is rich in high-quality,
general, semantically complete statements.

\begin{figure}[t]
\centerline{
 \fbox{%
   \parbox{1\columnwidth}{
     {\small
       \underline{\bf 1. Example generics about ``tree'' in \GenericsKB} \\
\blue{\bf Trees} are perennial plants that have long woody trunks. \\
\blue{\bf Trees} are woody plants which continue growing until they die. \\
Most \blue{\bf trees} add one new ring for each year of growth. \\
\blue{\bf Trees} produce oxygen by absorbing carbon dioxide from the air. \\
\blue{\bf Trees} are large, generally single-stemmed, woody plants. \\
\blue{\bf Trees} live in cavities or hollows. \\
\blue{\bf Trees} grow using photosynthesis, absorbing carbon dioxide and releasing oxygen. \\
\ \\
    \underline{\bf 2. An example entry, including metadata} \\
{\bf Term:} tree \\
{\bf Sent:} Most trees add one new ring for each year of growth. \\
{\bf Quantifier:} Most \\
{\bf Score:} 0.35 
\begin{des}
\item[{\bf Before:}] ...Notice how the extractor holds the core as it is removed from inside the hollow center of the bit. Tree cores are extracted with an increment borer. 
\item[{\bf After:}] The width of each annual ring may be a reflection of forest stand dynamics. Dendrochronology, the study of annual growth rings, has become prominent in ecology...
\end{des}
}}}
} 
\caption{Example generic statements in \GenericsKB, plus one showing associated metadata. \label{example}}
\end{figure}

Statements in \GenericsKB~were culled from over 1.7 billion sentences
from three corpora. To collect statements, we first clean the source
data, then filter it using linguistic rules to identify likely generics,
then apply a BERT-based scoring step to distinguish generics that
are meaningful on their own (avoiding generics with contextual meaning such
as {\it Meals are on the third floor}). The resulting KB
contains over 3.5M statements, each including metadata about
its topic, surrounding context, and a confidence measure.
Figure~\ref{example} illustrates some examples, as well
as a full entry illustrating the metadata.
We also create \GenericsKBBest~(1M+ sentences), containing the best-quality
generics in \GenericsKB~plus selected, synthesized generics
from WordNet and ConceptNet.


We also report results using \GenericsKB~for two tasks,
namely question-answering (using the OpenbookQA dataset \cite{Mihaylov2018CanAS}),
and explanation generation (using the QASC dataset \cite{khot2019qasc}).
Our goal is not to build a new model, but to see how an
existing model's performance changes when the \GenericsKB~corpus
replaces a larger corpus for these tasks.
We find that \GenericsKB~can sometimes produce higher
question-answering scores, and always produced
better quality explanations. This suggests that
\GenericsKB~may have value for other NLP tasks also, either standalone or
as an additional source of general knowledge to
help train models. Finally, independent of deep
learning, \GenericsKB~may be a valuable
resource for those studying generics and their
semantics in linguistics.

\section{Related Work}

A generic statement is one that makes a blanket statement
about the members of a category, e.g., ``Tigers are striped.''\footnote{
  We also include near-universally quantified statements
  such as ``Most tigers are striped'' in \GenericsKB,
  although their status as generics is sometimes disputed
  by semanticists.}
Because they apply to many entities, they are particularly
important for reasoning. Although common in language,
their semantics has been a topic of considerable debate in
linguistics, e.g., \cite{carlson1995generic,schubert1989generically,leslie2015generics,liebesman2011simple,schubert1987problems,generics-leslie}.
Rather than repeat that debate here, we note that our
primary goal is to {\it collect} rather than {\it interpret} generics.
We hope that our resource can contribute to study of their semantics.

Several repositories of general knowledge are available
already, but with different characteristics and coverage
to \GenericsKB, e.g., \cite{Sap2019ATOMICAA,tandon2014webchild,van2009deriving}.
ConceptNet \cite{speer2017conceptnet} is perhaps the
most used, containing approximately 1M English triples
(excluding RelatedTo, Synonym, and [Lexical]FormOf links),
or 34M triples total.
ConceptNet triples can be rendered as short generics,
thus covering just simple (typically three word) generic
statements about 28 relationships. Similarly, 
WordNet taxonomic and meronymic links express
short, specific relationships but leave most
uncovered (compare with Figure~\ref{example}).
Triple stores, e.g., \cite{clark2009large},
acquired from open information extraction \cite{banko2007open},
contain larger and less constrained collections of knowledge,
but typically with low precision \cite{Mishra2017DomainTargetedHP},
making it difficult to exploit them in practice. \GenericsKB~thus
fills a gap in this space, containing {\it naturally occurring}
generic statements that an author considered salient
enough to write down.

\section{Approach}

To construct \GenericsKB, sentences were selected from over 1.7B sentences in three corpora (Table~\ref{source-corpora}):
The Waterloo corpus is 280GB of English plain text, gathered by Charles Clarke (Univ. Waterloo) using a webcrawler in 2001
from .edu domains. It was made available to us and was previously used in \cite{clark2016combining}.
SimpleWikipedia is a filtered scrape of SimpleWikipedia pages (simple.wikipedia.org). 
The ARC corpus is a collection of 14M science and general sentences, released as part of the ARC challenge \cite{Clark2018ThinkYH}.
\GenericsKB~was then assembled in the following three steps:



 \eat{
   \begin{table}
    \setlength{\tabcolsep}{3pt}	
   \begin{tabular}
     {|l|rr|} \hline
     		& \multicolumn{2}{|c|}{Size} \\
     Corpus	& GB & \#Sent \\ \hline
     Waterloo	& 196 & 1700M \\
     SimpleWikipedia & 0.8 & 0.9M \\
     ARC	& 1.4 & 14M   \\ \hline
   \end{tabular}
   \caption{Source corpora used. \label{source-corpora}}
   \end{table}
   }

 \begin{table}
   \centerline{
     {\small
   \begin{tabular}
     {|l|rrr|} \hline
     & \multicolumn{3}{|c|}{\bf Size (\# sentences)} \\
     {\bf Corpus}	& {\bf Original} & {\bf Cleaned} & {\bf Filtered} \\ \hline
     Waterloo	& $\sim$ 1.7B & $\sim$ 500M & $\sim$ 3.1M \\
     SimpleWiki & $\sim$ 900k & $\sim$ 790k & $\sim$ 13k  \\
     ARC	& $\sim$  14M & $\sim$ 6.2M & $\sim$ 338k  \\ \hline
     {\bf \GenericsKB}	& $\sim$ 1.7B & $\sim$ 513M & $\sim$ 3.4M \\ \hline
   \end{tabular}
   }}
   \caption{Corpus sizes at different steps of processing. \label{source-corpora}}
   \end{table}

 \eat{
 \begin{table}
     {\small
   \begin{tabular}
     {|l|rrrr|} \hline
     Corpus	& Original & Cleaned & Filtered & Class- \\ 
		&	   &         &          & ified  \\ \hline
     Waterloo	& $\sim$ 1.7B & $\sim$ 500M & $\sim$ 3.6M & $\sim$ 2.9M \\
     SimpleWiki & $\sim$ 900k & $\sim$ 790k & $\sim$ 13k  & $\sim$ 13k \\
     ARC	& $\sim$  14M & $\sim$ 6.2M & $\sim$ 338k  & $\sim$ 330k \\ \hline
     \GenericsKB	& $\sim$ 1.7B & $\sim$ 513M & $\sim$ 4M & $\sim$ 3.4M \\ \hline
   \end{tabular}
   }
   \caption{Resources sizes through the pipeline. \label{source-corpora}}
   \end{table}
}

 \subsection{Cleaning}

 As the source corpora originated from web scrapes, they contain noise in various forms, such as blocks of code, non-English text, hyperlinks, and emails. The corpora were cleaned using the following:
 \vspace{-2mm}
\begin{ite}
\item Regular Expressions to capture frequently occurring lexical properties of noise. 
\item Sentence and token length heuristics to filter out malformed sentences.
\item Text cleanup using the Fixes Text For You (ftfy) python library which fixes various encoding-related errors.
\item Language Detection using spaCy to filter out non-English text.
\end{ite}

\subsection{Filtering}

We next use a set of 27 hand-authored lexico-syntactic rules 
to identify standalone generic sentences, and reject others. For example,
sentences that start with a bare plural (``Dogs are...'') are considered good candidates,
while those starting with a determiner (``A man said...'') or containing
a present participle (``A bear is running...'') are not. Similarly,
sentences containing pronouns (``He said...'') are likely to have contextual
rather than standalone meaning, and so are also rejected. A sample of
the filtering rules are summarized in Figure~\ref{rules}, and 
the full list of rules is given in the Appendix. Given the size and redundancy of the
initial corpus, these rules aim to filter the corpus aggressively to produce
a set of high-quality candidates, rather than catch all possible standalone generics.

\begin{figure}[t]
\centerline{
 \fbox{%
   \parbox{1\columnwidth}{
     {\small
  {\bf no-bad-first-word:} Sentence does not start with a determiner (``a'', ``the'',...) or selected other words.\\
  {\bf remove-non-verb-roots:} Remove if root is a non-verb \\
  {\bf remove-present-participle-roots:} Do not consider any present participle roots. \\  
  {\bf has-no-modals:} Sentences containing modals (``could'', ``would'', etc) are rejected \\
  {\bf all-propn-exist-in-wordnet:} All (normalized, non-stop) words are in WordNet's vocabulary 
}}}
} 
\caption{Example filtering rules. (See supplementary material for the full list). \label{rules}}
\end{figure}

\subsection{Scoring \label{scoring}}

Finally, we train and apply a BERT classifier to score sentences by
by how well they describe a {\it useful, general truth}. To build
the classifier, a random subset (size 10k) of the 3.4M candidate generics
was labeled by crowdworkers as to whether they expressed a {\it useful, general truth} about
 the world (with options yes, no, unsure), guided by examples. Specifically, workers were
 asked to reject \\
 (1) sentences which do not stand on their own, e.g.,:
 \begin{myquote2}
   {\it Free parking is provided}
   \end{myquote2}
 (2) subjective and/or not useful statements, e.g.,
 \begin{myquote2}
   {\it Life is too serious, sometimes.}
   \end{myquote2}
 (3) Vague statements, e.g.,
 \begin{myquote2}
   {\it All cats are essentially cats.}
 \end{myquote2}   
  (4) Statements about people and companies, e.g.,
 \begin{myquote2}   
   {\it Apple makes lots of iPhones}
 \end{myquote2}
 (5) Facts that are incorrect in isolation, e.g.,
 \begin{myquote2}   
   {\it All maps are hand-drawn.}
 \end{myquote2}
 Each fact was annotated twice and scores (yes/unsure/no = 1/0.5/0) averaged. The
 joint probability of agreement (i.e., that both annotators agreed) was 70.1\% (approximately 1/3 of the agreed annotations being ``yes'', 2/3 ``no''), and Cohen's Kappa $\kappa$ was 0.52 (``moderate agreement'' ).
The dataset was then split 70:10:20 into train:dev:test, and a BERT classifer\footnote{We use
the BERT-for-classification package provided by AllenNLP,
https://allenai.github.io/allennlp-docs/api/allennlp.models.bert\_for\_classification.html}
fine-tuned on the training set. Each sentence is input simply as {\it [CLS] sentence}. The output is pooled,
then run through a linear layer which outputs two logits representing the two classes (yes/no),
followed by a softmax to obtain class probabilities. This classifier scored
83\% on the held-out test set. The classifier was then used to score all
3.4M extracted generic sentences.

\subsection{\GenericsKB~and \GenericsKBBest}

The final \GenericsKB~contains 3,433,000 sentences. We also create
\GenericsKBBest, comprising \GenericsKB~generics with a score $>$ 0.23\footnote{By calibration, equivalent to an annotator score of 0.5, i.e., more likely good than bad.},
augmented with short generics synthesized from three other resources\footnote{
  ConceptNet (isa, hasPart, locatedAt, usedFor); WordNet (isa, hasPart); and
  the Aristo TupleKB (at https://allenai.org/data/tuple-kb)
For WordNet, we use just the most frequent sense for each generic term.}
for all the terms (generic categories) in \GenericsKBBest.
\GenericsKBBest~contains 1,020,868 generics (774,621 from \GenericsKB~plus 246,247 synthesized).

\section{Evaluation}

For some initial indications of whether \GenericsKB~can be useful, we
performed two experiments.

\subsection{Question-Answering}

We evaluate using \GenericsKB~for a question-answering task, namely
OpenbookQA \cite{Mihaylov2018CanAS}, comparing it
to using an alternative, large, publically available corpus (QASC-17M, \cite{khot2019qasc}).
For both, we use the BERT-MCQ QA system \cite{khot2019qasc}.
Note that our goal is to evaluate the corpora, not the QA system.
The results are shown in Table~\ref{comparison}, indicating that
using the high-quality version \GenericsKBBest~can, at least in this case, result in improved
QA performance over using the original corpus, even though it is
a fraction of the size. 

\begin{table}
  \centerline{
  {\small
       \begin{tabular}
         {|c|c|c|} \hline
         {\bf Corpus} & {\bf Size} & {\bf Score on OBQA (test)} \\ \hline
         QASC-17M & 17M & 0.660 \\ 
         \GenericsKB  & 3.4M & 0.632 \\
         {\bf \GenericsKBBest} & 1M & {\bf 0.678} \\ \hline
       \end{tabular}
      }}
     \caption{Comparative performance of different corpora
       for answering OBQA questions. \label{comparison}}
 \end{table}

\begin{table}
  \centerline{
     {\small
       \begin{tabular}
         {|c|cc|} \hline
         & \multicolumn{2}{|c|}{\bf Explanation Quality} \\
         {\bf Corpus} & {\bf on OBQA} & {\bf on QASC} \\ \hline
         QASC-17M & 0.44 & 0.66 \\
         {\bf \GenericsKBBest} & {\bf 0.61} & {\bf 0.79} \\ \hline
       \end{tabular}
     }}
     \caption{Comparative quality of two-hop explanations (sentence chains),
       generated using two different corpora for two different question sets. \label{explanations}}
 \end{table}

 \subsection{Explanation Quality}

 We also experimented with using \GenericsKBBest~to generate
 {\it explanations} for a (given) answer, where an explanation
 is a chain of two sentences drawn from the corpus. 
 For example: \\
 {\it What can cause a forest fire? storms} {\bf because:} \\
\hspace*{14mm} {\it Storms can produce lightning} \\
\hspace*{5mm} AND {\it Lightning can start fires} \\
Good explanations typically use generic sentences, reflecting the underlying
formal structure of the explanation. This suggests that a
corpus of generics may help in this task.

We test this hypothesis using the QASC dataset. We can do
this because the BERT-MCQ system described earlier already
finds candidate good chains as part of its
retrieval step \cite{khot2019qasc} (specifically, it finds
pairs of sentences from the corpus that maximally overlap
the question, answer, and each other). We can thus collect
these chains found using the original QASC-17M corpus,
and using \GenericsKBBest, and compare quality.

To evaluate these chains, we train a simple BERT-model using the QASC
training data, which comes with a gold reasoning chain for every correct answer.
We use the gold chains as examples of good chains,
and BERT-MCQ-generated chains for incorrect answer options as examples of bad (invalid) chains.
We can then use the trained model to evaluate the chains collected earlier.

The results are in Table~\ref{explanations}, and indicate that substantially better explanations
are generated with \GenericsKBBest.
The same result was found using the OBQA dataset.
In particular, because of the eclectic nature of
the QASC-17M corpus, nonsensical explanations can
often occur, e.g.,:\\
{\it What do vehicles transport? people} {\bf because:} \\
\hspace*{14mm} {\it What to say what vehicle to use} \\
\hspace*{5mm} AND {\it Now people say it's time to move on.}  \\
compared with the \GenericsKBBest~explanation: \\
{\it What do vehicles transport? people} {\bf because:} \\
\hspace*{14mm} {\it A vehicle is transport} \\
\hspace*{5mm} AND {\it Transportation is used for moving people}\\
Here, the QASC-17M explanation is nonsensical, while as \GenericsKB~is rich in
stand-alone generics, the explanations produced with it are more often valid.

\eat{
  \subsection{Experiments with ARC}

Finally, we present a negative but informative result.
We compared using \GenericsKBBest~with using the Waterloo corpus (Table~\ref{source-corpora})
for the well-known ARC QA dataset \cite{Clark2018ThinkYH}, using a BERT-based QA system\footnote{
  We follow the implementation described at leaderboard.allenai.org/arc/submission/blcotvl7rrltlue6bsv0}.
The results (Table~\ref{arc}) show that \GenericsKBBest~cannot replace the Waterloo corpus
without loss of performance. This is perhaps not surprising given their relative sizes
(1 million vs. 1.7 billion sentences). It illustrates that although \GenericsKB~contains substantial
general knowledge, it is not a full enumeration of all that is needed to answer
complex questions. Using sentences from both together (using 8 from Waterloo, 2 \GenericsKB,
last column in Table~\ref{arc}) did not significantly affect results.

\begin{table}
  {\small
    \centerline{
   \setlength{\tabcolsep}{2pt} 
   \begin{tabular}{|c|c|ccc|} \hline
        &   & \multicolumn{3}{|c|}{\bf Corpus} \\
    {\bf Test Set} & {\bf Num Q} & {\bf Waterloo} & {\bf \GenericsKBBest} & {\bf Both (4:1)}  \\ \hline
    ARC-Easy & 2375 & 82.82 & 73.73 & 82.15 \\
    ARC-Chal & 1172 & 66.38 & 58.96 & 65.44 \\ \hline
   \end{tabular}
  }}
  \caption{Scores using different corpora on ARC. \label{arc}}
\end{table}
}


\subsection{\GenericsKB~Quality}
\vspace{-1mm}

Finally we note that even with filtering, some (undesirable) contextual
generics occasionally pass through. Examples include:
\vspace{-2mm}
\begin{ite}
\item {\it All results are confidential.} 
\item {\it Complications are usually infrequent.} 
\item {\it Democracy is four wolves and a lamb voting on what to have for lunch.}
\end{ite}
\vspace{-2mm}
These examples exhibit ellipsis, vagueness, and metaphor,
complicating their interpretation. Ideally, the scoring model
would then score these low, 
but this may not always happen: recognizing contextuality often
requires world knowledge. For example, consider distinguishing the
good, standalone generic {\it Murder is illegal} from the contextual one {\it Parking is illegal}.

To evaluate the extent of this, two annotators independently
annotated 100 random (\GenericsKB) sentences from \GenericsKBBest~as
to whether they represented {\it useful, general truths}
(the same criterion as in Section~\ref{scoring}), and
found 85\% (averaged) met this criterion. This suggests
that such problems are relatively uncommon.


\section{Conclusion}
With the growing use of deep learning in NLP, researchers
have often sought out additional general knowledge resources
to improve their systems. To help meet this need, as well
as provide a general resource for linguistics, we have
created \GenericsKB, the first large-scale resource of
{\it naturally occurring} generic statements,
as well as an augmented subset \GenericsKBBest, including
important metadata about each statement.
While \GenericsKB~is not a replacement for a Web-scale corpus,
we have shown it can assist in both question-answering
and explanation construction for two existing datasets.
These positive examples of utility suggest that \GenericsKB~has
potential as a large, new resource of general
knowledge for the community. \GenericsKB~is available
at https://allenai.org/data/genericskb.

%% file: appendix.tex
\clearpage

\section*{Appendix: Patterns for Identifying Generics}

The following 27 rules are used to identify generic sentences, as well as help filter out those which are likely contextual, gibberish,
or otherwise not stand-alone. Some rules use spaCy features for processing. To be retained, each sentence must pass the following tests:
\begin{myquote}
{\bf is-short-enough:} Length of the sentence $\leq$ 100. \\
{\bf starts-with-capital:} The first character is an upper-case character. \\
{\bf ends-with-period:} The last character is a period. \\
{\bf has-at-least-one-token:} The sentence contains at least one spaCy token. \\
{\bf has-no-bad-first-word:} The first word is not in a list of bad-first-words (determiners, etc.) \\
{\bf has-no-bad-words:} The sentence does not contain words in a badword list (e.g., copyright, licence, ...) \\
{\bf has-no-bad-pronouns:} The sentence does not contain personal pronouns (he, she, ...) \\
{\bf has-no-negations:} The sentence does not contain negations. \\
{\bf has-no-modals:} The sentence does not contain modals (``would'', ``should'',...).  \\
{\bf first-word-is-not-verb:} The first word of the sentence is not a verb. \\
{\bf first-word-is-not-conjunction:} The first word is not a conjunction. \\
{\bf look-for-positive-quantifier-at-first-word:} If the first word is a positive quantifier (``all'', ``some''), note the quantifier and repeat the filter using the sentence without the quantifier. \\
{\bf has-acceptable-past-participle-root:} The root verb is in the present passive, or is not a past participle. \\
{\bf noun-exists-before-root:} There is a 'NOUN' token before the root. \\
{\bf key-concept-head-pos-tags-not-contradicted-by-wordnet:} If WordNet disagrees about the POS of the key concept head, filter out this sentence. \\
{\bf has-no-digits:} The sentence has no digits. \\
{\bf all-propn-exist-in-wordnet:} All PROPN tokens exist in WordNet. \\
{\bf all-propn-have-acceptable-ne-labels:} Any PROPN tokens have one of the following ent\_type values: 'EVENT', 'GPE', 'LANGUAGE', 'LAW', 'LOC', 'WORK\_OF\_ART'. (These acceptable values were decided by the corresponding top level rules.)
\end{myquote}
and must \underline{not} pass these tests:
\begin{myquote}
{\bf scr.dot\_dot\_in\_sentence:} There is '..' in the sentence. \\
{\bf scr.www\_in\_sentence:} There is 'www' in the sentence. \\
{\bf scr.com\_in\_sentence:} There is '.com' in the sentence. \\
{\bf scr.many\_hyphens\_in\_sentence:} The number of hyphens in the sentence is $\geq$ 2. \\
{\bf scr.sentence\_does\_not\_end\_with\_period:} The sentence does not end with a period. \\
{\bf remove-non-verb-roots:} Remove any sentences with non-verbal roots (e.g., ``A large tree.''). \\
{\bf remove-present-participle-roots:} Reject sentences whose root verb is a present participle (``sitting'',...). \\
{\bf remove-first-word-roots:} Reject sentences with a root that corresponds to the first word. \\
{\bf remove-past-tense-roots:} Reject sentences with any past tense roots (``ate'',...). 
\end{myquote}

%% file: main.bbl
\begin{thebibliography}{24}
\expandafter\ifx\csname natexlab\endcsname\relax\def\natexlab#1{#1}\fi

\bibitem[{Banko et~al.(2007)Banko, Cafarella, Soderland, Broadhead, and
  Etzioni}]{banko2007open}
Michele Banko, Michael~J Cafarella, Stephen Soderland, Matthew Broadhead, and
  Oren Etzioni. 2007.
\newblock Open information extraction from the web.
\newblock In \emph{Proc. IJCAI'07}, volume~7, pages 2670--2676.

\bibitem[{Carlson and Pelletier(1995)}]{carlson1995generic}
Gregory~N Carlson and Francis~Jeffry Pelletier. 1995.
\newblock \emph{The generic book}.
\newblock University of Chicago Press.

\bibitem[{Clark et~al.(2018)Clark, Cowhey, Etzioni, Khot, Sabharwal, Schoenick,
  and Tafjord}]{Clark2018ThinkYH}
Peter Clark, Isaac Cowhey, Oren Etzioni, Tushar Khot, Ashish Sabharwal, Carissa
  Schoenick, and Oyvind Tafjord. 2018.
\newblock Think you have solved question answering? {T}ry {ARC}, the {AI2}
  {R}easoning {C}hallenge.
\newblock \emph{ArXiv}, abs/1803.05457.

\bibitem[{Clark et~al.(2016)Clark, Etzioni, Khot, Sabharwal, Tafjord, Turney,
  and Khashabi}]{clark2016combining}
Peter Clark, Oren Etzioni, Tushar Khot, Ashish Sabharwal, Oyvind Tafjord,
  Peter~D Turney, and Daniel Khashabi. 2016.
\newblock Combining retrieval, statistics, and inference to answer elementary
  science questions.
\newblock In \emph{AAAI}, pages 2580--2586.

\bibitem[{Clark and Harrison(2009)}]{clark2009large}
Peter Clark and Phil Harrison. 2009.
\newblock Large-scale extraction and use of knowledge from text.
\newblock In \emph{Proceedings of the fifth international conference on
  Knowledge capture}, pages 153--160. ACM.

\bibitem[{Fellbaum(1998)}]{fellbaum1998wordnet}
Christiane Fellbaum. 1998.
\newblock \emph{WordNet}.
\newblock Wiley Online Library.

\bibitem[{Icarte et~al.(2017)Icarte, Baier, Ruz, and Soto}]{Icarte2017HowAG}
Rodrigo~Toro Icarte, Jorge~A. Baier, Cristian Ruz, and Alvaro Soto. 2017.
\newblock How a general-purpose commonsense ontology can improve performance of
  learning-based image retrieval.
\newblock In \emph{IJCAI}.

\bibitem[{Khot et~al.(2019)Khot, Clark, Guerquin, Jansen, and
  Sabharwal}]{khot2019qasc}
Tushar Khot, Peter Clark, Michal Guerquin, Peter Jansen, and Ashish Sabharwal.
  2019.
\newblock Qasc: A dataset for question answering via sentence composition.
\newblock \emph{arXiv preprint arXiv:1910.11473}.
\newblock (AAAI'20, to appear).

\bibitem[{Leslie(2011)}]{generics-leslie}
Sarah-Jane Leslie. 2011.
\newblock Generics.
\newblock In \emph{The Routledge Encyclopedia of Philosophy}. Routledge.

\bibitem[{Leslie(2015)}]{leslie2015generics}
Sarah-Jane Leslie. 2015.
\newblock Generics oversimplified.
\newblock \emph{No{\^u}s}, 49(1):28--54.

\bibitem[{Liebesman(2011)}]{liebesman2011simple}
David Liebesman. 2011.
\newblock Simple generics.
\newblock \emph{No{\^u}s}, 45(3):409--442.

\bibitem[{Liu et~al.(2019)Liu, Zhou, Zhao, Wang, Ju, Deng, and
  Wang}]{Liu2019KBERTEL}
Weijie Liu, Peng Zhou, Zhe Zhao, Zhiruo Wang, Qi~Ju, Haotang Deng, and Ping
  Wang. 2019.
\newblock K-bert: Enabling language representation with knowledge graph.
\newblock \emph{ArXiv}, abs/1909.07606.

\bibitem[{Mihaylov et~al.(2018)Mihaylov, Clark, Khot, and
  Sabharwal}]{Mihaylov2018CanAS}
Tzvetan Mihaylov, Peter~F. Clark, Tushar Khot, and Ashish Sabharwal. 2018.
\newblock Can a suit of armor conduct electricity? a new dataset for open book
  question answering.
\newblock In \emph{EMNLP}.

\bibitem[{Mishra et~al.(2017)Mishra, Tandon, and
  Clark}]{Mishra2017DomainTargetedHP}
Bhavana~Dalvi Mishra, Niket Tandon, and Peter Clark. 2017.
\newblock Domain-targeted, high precision knowledge extraction.
\newblock \emph{Transactions of the Association for Computational Linguistics},
  5:233--246.

\bibitem[{Paul and Frank(2019)}]{Paul2019RankingAS}
Debjit Paul and Anette Frank. 2019.
\newblock Ranking and selecting multi-hop knowledge paths to better predict
  human needs.
\newblock In \emph{NAACL'19}.

\bibitem[{Peters et~al.(2019)Peters, Neumann, RobertLLogan, Schwartz, Joshi,
  Singh, and Smith}]{Peters2019KnowledgeEC}
Matthew~E. Peters, Mark Neumann, IV~RobertLLogan, Roy Schwartz, Vidur Joshi,
  Sameer Singh, and Noah~A. Smith. 2019.
\newblock Knowledge enhanced contextual word representations.
\newblock In \emph{EMNLP}.

\bibitem[{Sap et~al.(2019)Sap, Bras, Allaway, Bhagavatula, Lourie, Rashkin,
  Roof, Smith, and Choi}]{Sap2019ATOMICAA}
Maarten Sap, Ronan~Le Bras, Emily Allaway, Chandra Bhagavatula, Nicholas
  Lourie, Hannah Rashkin, Brendan Roof, Noah~A. Smith, and Yejin Choi. 2019.
\newblock Atomic: An atlas of machine commonsense for if-then reasoning.
\newblock In \emph{AAAI}.

\bibitem[{Schubert and Pelletier(1987)}]{schubert1987problems}
Lenhart~K Schubert and Francis~J Pelletier. 1987.
\newblock Problems in the representation of the logical form of generics,
  plurals, and mass nouns.
\newblock \emph{New directions in semantics}, pages 385--451.

\bibitem[{Schubert and Pelletier(1989)}]{schubert1989generically}
Lenhart~K Schubert and Francis~Jeffry Pelletier. 1989.
\newblock Generically speaking, or, using discourse representation theory to
  interpret generics.
\newblock In \emph{Properties, types and meaning}, pages 193--268. Springer.

\bibitem[{Speer et~al.(2017)Speer, Chin, and Havasi}]{speer2017conceptnet}
Robyn Speer, Joshua Chin, and Catherine Havasi. 2017.
\newblock Conceptnet 5.5: An open multilingual graph of general knowledge.
\newblock In \emph{AAAI}.

\bibitem[{Tandon et~al.(2014)Tandon, De~Melo, Suchanek, and
  Weikum}]{tandon2014webchild}
Niket Tandon, Gerard De~Melo, Fabian Suchanek, and Gerhard Weikum. 2014.
\newblock Webchild: Harvesting and organizing commonsense knowledge from the
  web.
\newblock In \emph{Proceedings of the 7th ACM international conference on Web
  search and data mining}, pages 523--532. ACM.

\bibitem[{Van~Durme et~al.(2009)Van~Durme, Michalak, and
  Schubert}]{van2009deriving}
Benjamin Van~Durme, Phillip Michalak, and Lenhart~K Schubert. 2009.
\newblock Deriving generalized knowledge from corpora using wordnet
  abstraction.
\newblock In \emph{Proceedings of the 12th Conference of the European Chapter
  of the Association for Computational Linguistics}, pages 808--816.
  Association for Computational Linguistics.

\bibitem[{Wang et~al.(2018)Wang, Durrett, and Erk}]{Wang2018ModelingSP}
Su~Wang, Greg Durrett, and Katrin Erk. 2018.
\newblock Modeling semantic plausibility by injecting world knowledge.
\newblock In \emph{NAACL-HLT}.

\bibitem[{Yang et~al.(2019)Yang, Wang, Liu, Liu, Lyu, Wu, She, and
  Li}]{Yang2019EnhancingPL}
An~Yang, Quan Wang, Jing Liu, Kai Liu, Yajuan Lyu, Hua Wu, Qiaoqiao She, and
  Sujian Li. 2019.
\newblock Enhancing pre-trained language representations with rich knowledge
  for machine reading comprehension.
\newblock In \emph{ACL}.

\end{thebibliography}
